\documentclass[acmtog]{acmart}
\acmSubmissionID{921}
\usepackage{algorithm}
\usepackage{algorithmic}
\usepackage{sidecap}

\begin{document}

\title{PDT: Point Distribution Transformation with Diffusion Models}

\author{Jionghao Wang}
\authornote{Both authors contributed equally to this research.}
\email{jionghao@tamu.edu}
\affiliation{%
  \institution{Texas A\&M University}
  \country{USA}}
\orcid{0009-0002-9683-8547}

\author{Cheng Lin}
\authornotemark[1]
\authornote{Corresponding authors.}
\email{chlin@connect.hku.hk}
\affiliation{%
  \institution{The University of Hong Kong}
  \country{China}
}
\orcid{0000-0002-3335-6623}

\author{Yuan Liu}
\affiliation{%
  \institution{The Hong Kong University of Science and Technology}
  \country{China}}
\email{liuyuanwhuer@gmail.com}
\orcid{0000-0003-2933-5667}

\author{Rui Xu}
\email{ruixu1999@connect.hku.hk}
\affiliation{%
  \institution{The University of Hong Kong}
  \country{China}
}
\orcid{0000-0001-8273-1808}

\author{Zhiyang Dou}
\email{zhiyang0@connect.hku.hk}
\affiliation{%
  \institution{The University of Hong Kong}
  \country{China}
}
\orcid{0000-0003-0186-8269}

\author{Xiao-xiao Long}
\affiliation{%
  \institution{Nanjing University}
  \country{China}}
\email{xxlong@connect.hku.hk}
\orcid{0000-0002-3386-8805}

\author{Hao-xiang Guo}
\affiliation{%
  \institution{Skywork AI, Kunlun Inc.}
  \country{China}}
\email{guohaoxiangxiang@gmail.com}
\orcid{0009-0009-0002-5252}

\author{Taku Komura}
\affiliation{%
  \institution{The University of Hong Kong}
  \country{China}
}
\email{taku@cs.hku.hk}
\orcid{0000-0002-2729-5860}

\author{Wenping Wang}
\email{wenping@tamu.edu}
\orcid{0000-0002-2284-3952}
\author{Xin Li}
\authornotemark[2]
\email{xinli@tamu.edu}
\affiliation{%
  \institution{Texas A\&M University}
  \country{USA}}
\orcid{0000-0002-0144-9489}

\renewcommand{\shortauthors}{Wang, Lin et al.}

\begin{abstract}
  Point-based representations have consistently played a vital role in geometric data structures. Most point cloud learning and processing methods typically leverage the unordered and unconstrained nature to represent the underlying geometry of 3D shapes. However, how to extract meaningful structural information from unstructured point cloud distributions and transform them into semantically meaningful point distributions remains an under-explored problem.
  We present PDT, a novel framework for point distribution transformation with diffusion models. Given a set of input points, PDT learns to transform the point set from its original geometric distribution into a target distribution that is semantically meaningful. Our method utilizes diffusion models with novel architecture and learning strategy, which effectively correlates the source and the target distribution through a denoising process. Through extensive experiments, we show that our method successfully transforms input point clouds into various forms of structured outputs - ranging from surface-aligned keypoints, and inner sparse joints to continuous feature lines. The results showcase our framework's ability to capture both geometric and semantic features, offering a powerful tool for various 3D geometry processing tasks where structured point distributions are desired. Code will be available at this link: \href{https://github.com/shanemankiw/PDT}{link}.
\end{abstract}



\begin{teaserfigure}
  \centering
  \includegraphics[width=0.95\textwidth]{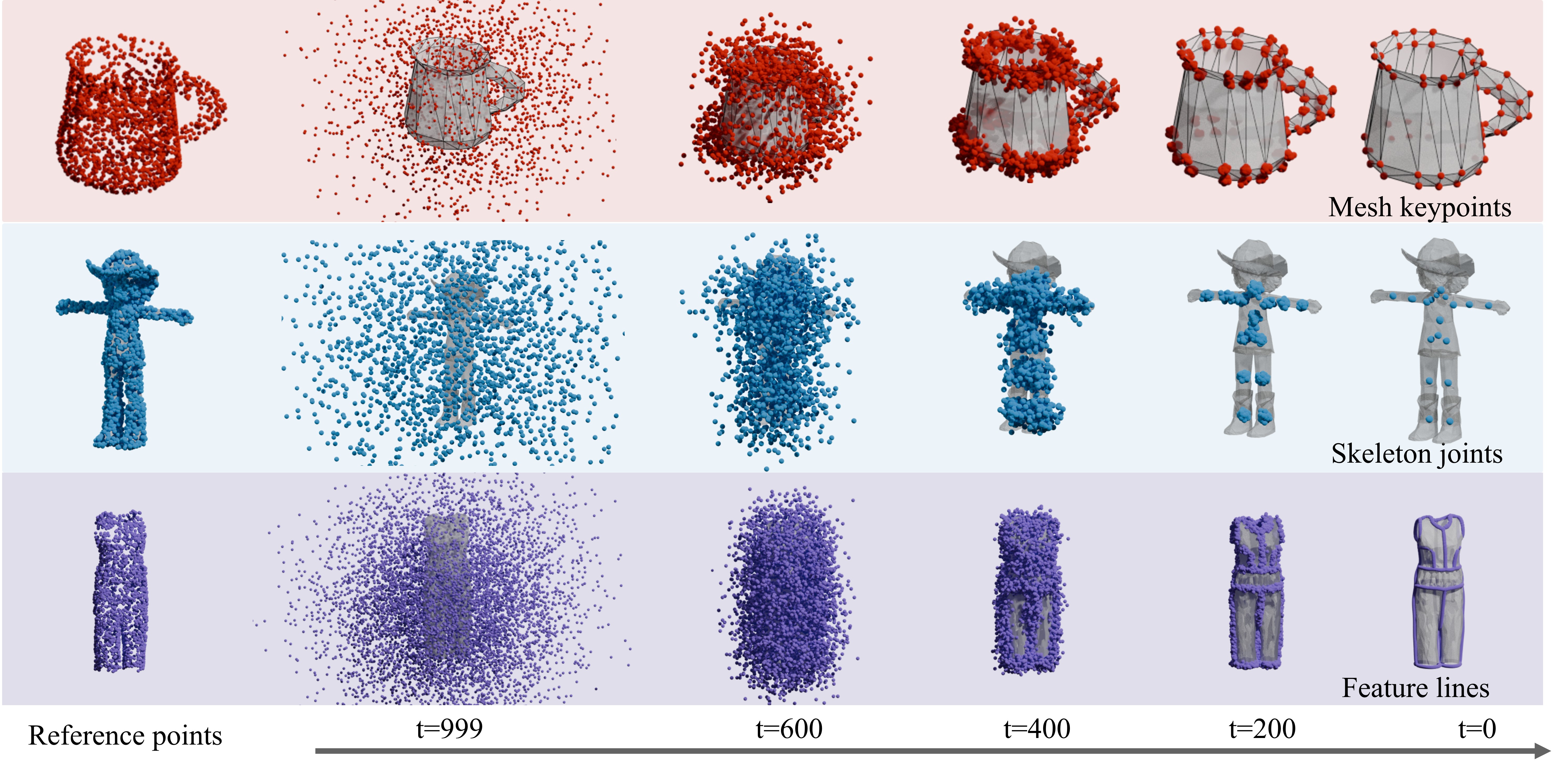}
  \caption{Overview of PDT. PDT leverages a diffusion transformer-based architecture to transform Gaussian noise into semantically meaningful point distributions, guided by input reference points. We demonstrate the effectiveness of our approach across three structural representations: surface keypoints for artist-inspired meshes, inner skeletal joints for character rigging, and continuous feature lines for garment analysis. Note that meshes in this figure are only for reference.}
  \label{fig:teaser}
\end{teaserfigure}

\maketitle

\section{Introduction}
\label{sec:intro}
Point-based 3D shape representations play a crucial role in characterizing geometry distributions. The flexibility, simplicity, and scalability of points make point clouds a fundamental component of geometric data structures. However, due to the unstructured and irregular nature of point clouds, effectively perceiving high-level semantic information from irregular point distributions and generating structurally meaningful representations remains a challenging task.

Existing methods often attempt to encode dense point clouds into implicit global features, followed by direct regression or classification to structured outputs (e.g., primitives, meshes, or skeletons). These approaches typically rely on deterministic feedforward mappings, which may struggle to capture the multimodal nature of semantic structures or maintain local geometric fidelity. In contrast, we propose a generative formulation that treats both the input and target structures as point distributions and learns to transform one into the other via guided denoising. This distribution-to-distribution mapping enables explicit pointwise correspondence, better handling structural ambiguity and producing compact, semantically meaningful outputs. As recent advances~\cite{he2024lotus, fu2024geowizard} demonstrate, generative models offer greater flexibility and robustness for geometry reasoning tasks. While diffusion models have shown strong performance in 3D generation, their potential for transforming point distributions into structured representations remains underexplored—this work aims to fill that gap. 

In this paper, we introduce a novel framework, named PDT, for point distribution transformation using diffusion models. Given a set of input points representing the shape geometry, our method learns to transform the points from their original surface distribution, where the input point cloud is viewed as samples drawn from the underlying surface probability distribution, into a semantically meaningful target distribution. The target structured points, such as keypoints or feature lines, are treated as a separate probability distribution in 3D space.
We establish a probabilistic mapping between the two distributions by injecting Gaussian noise to the input points and learning to progressively denoise them toward the target points, correlating the surface geometry distribution and the target structure distribution through explicit per-point guidance. 

Our method is simple yet effective overall, with several key designs that make it well-suited for transformation between different point distributions. First, by introducing explicit correspondences between each noisy input point and a target structure point, we enable the denoising process to effectively perceive local geometry in specific regions, thereby generating target points that align with the local structures. Second, since the distribution of semantically structured points is typically compact and sparse, while the input point distribution is usually dense, our method should accurately map each point from an excessive point set to a precise location within a sparse set of target points. To enhance the denoising process's ability to capture fine-grained details, we carefully design a noise schedule that strengthens the learning of local details. Additionally, we introduce a gradient-guided strategy during inference, which makes the point distribution transformation more controllable and precise.

We demonstrate our method in three tasks where the input points are transformed into different semantically meaningful target distributions reflecting distinct structural priors: 1) surface-aligned mesh keypoints; 2) inner skeletal joints; and 3) continuous feature lines. These three distributions reflect different structural dependencies between the target and the source point geometry. Through extensive experiments, we validate the strong structural prediction and perception capabilities of our method across these tasks, demonstrating its ability to effectively generate semantically meaningful point distributions. This, in turn, facilitates improvements in corresponding downstream tasks. Our results highlight that the proposed method serves as a general-purpose framework for point distribution transformation, exhibiting remarkable diversity, versatility, and potential for further extension.

\section{Related Work}
\label{sec:literature}

\begin{figure*}
    \centering
    \includegraphics[width=\linewidth]{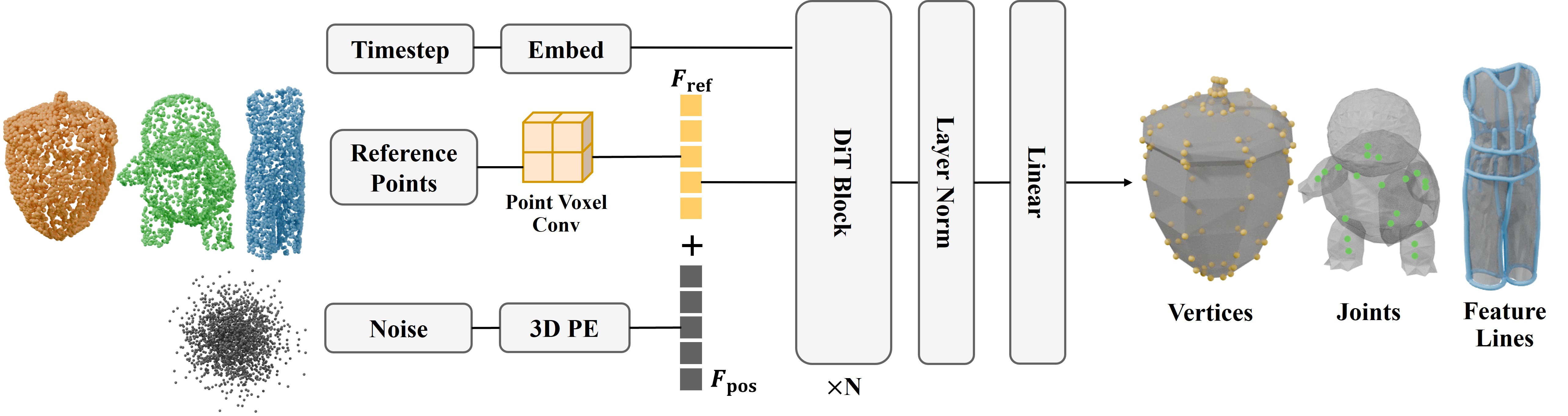}
    \caption{Architecture overview of our PDT. The model extracts per-point features $\mathbf{F}_{\text{ref}}$ with Point Voxel Conv(PVCNN)~\cite{liu2019point} from input reference points and associates them with corresponding noisy points through adding its positional encoding features $\mathbf{F}_{\text{pos}}$ through a learnable positional embedding(3D PE). The combined features and timestep embeddings are processed through a series of DiT layers to learn the distribution transformation.}
    \label{fig:model}
\end{figure*}

\subsection{Learning Point-based Transformations}
Point-based transformation methods have been an active area of research in 3D geometry processing. Early approaches focused on specific geometric transformations, such as surface-to-skeleton mapping~\cite{lin2021point2skeleton, yin2018p2p} and point cloud completion~\cite{wen2021pmp}. While these methods demonstrate effectiveness in their targeted domains, their reliance on deterministic feed-forward networks limits their capability to model complex distribution patterns and capture intricate geometric relationships. While methods such as VoteNet~\cite{qi2019deep} and Robust Symmetry Detection via Riemannian Langevin Dynamics~\cite{je2024robust} excel at extracting sparse structures for specific tasks like object and symmetry detection, our approach offers a more flexible pipeline. Through task-specific training, it can effectively address objectives ranging from highly sparse to relatively dense, unlike methods inherently tailored to sparse outputs.

Recent advances in denoising diffusion probabilistic models (DDPM) \cite{ho2020denoising} and their variants~\cite{peebles2023scalable, nichol2021improved} have shown remarkable success in various 3D generation tasks. These approaches have been effectively applied to diverse 3D representations, including 3D Gaussians~\cite{tang2025lgm}, neural fields~\cite{zhang20233dshape2vecset, zhang2024clay, yu2023surf}, and multi-view images~\cite{long2024wonder3d, li2024era3d}. In the context of point clouds specifically, several works have leveraged diffusion models for generation tasks, such as PVD~\cite{zhou20213d}, Point-E~\cite{nichol2022point}, and LION~\cite{vahdat2022lion}. Notably, DiT-3D~\cite{mo2023dit} pioneered the integration of the DiT architecture~\cite{peebles2023scalable} into point cloud processing.

Of particular relevance to our work is Geometry Distribution~\cite{biao2024geometry}, which demonstrates the capability of diffusion models to learn and represent surface point distributions. While they focus on modeling individual distributions as shape representations, our work extends this concept by learning the transformations between different point distributions. 

\subsection{Structured Representations}
We examine three fundamental types of structured point-based representations: surface keypoints, skeletal joints, and garment feature lines, each serving distinct purposes in 3D geometry processing and computer graphics applications.

\subsubsection{Surface keypoints and remeshing.}
Surface keypoints serve as critical geometric landmarks that capture the essential structural and semantic features of 3D shapes. The generation of semantically meaningful surface keypoints represents a fundamental challenge in geometric processing, particularly for creating artist-inspired meshes. 
Traditional remeshing approaches, including Quadric Error Metrics (QEM)~\cite{garland1997surface,ozaki2015out,wei2010feature,panchal2022feature,liu2023surface, chen2023robust} and Centroidal Voronoi Tessellation (CVT)~\cite{du1999centroidal,sun2011obtuse,edwards2013kCVT,CVTyan2015non,du2018field,CVTwang2018isotropic,levy2010p}, primarily focus on geometric feature preservation through optimization-based techniques. While these methods excel at generating isotropic meshes, they often fail to capture the semantic understanding inherent in artist-created representations.

Recently, a series of methods for compact mesh generation have demonstrated impressive results~\cite{achiam2023gpt, wang2024llama, weng2024pivotmesh, 
  chen2024meshanything, chen2024meshxl}. These methods predominantly operate by treating mesh faces as token sequences and generating them through autoregressive processes. Their share a similar idea of treating mesh faces as token sequences and generating them autoregressively. In contrast, we ventured down a completely new and distinct path by directly generating the 3D guiding keypoints that serve as a foundation followed by a simple remeshing process.

\subsubsection{Skeletal joints.}
Automatic rigging is an important task and can reduce the manual effort required from artists in character creation and animation processes. Skeletal joints serve as fundamental control points for character rigging, enabling subsequent animation capabilities~\cite{liao2022skeleton}. Traditional approaches, such as Pinocchio~\cite{baran2007automatic}, rely on fixed templates, limiting their ability to accommodate diverse skeletal structures. While learning-based methods such as RigNet~\cite{xu2020rignet} have demonstrated promising results in joint prediction, their reliance on deterministic feed-forward networks constrains their ability to model complex skeletal distribution patterns and capture intricate anatomical relationships. In contrast, our PDT framework leveraging diffusion models demonstrates superior precision and robustness.

\subsubsection{Garment feature lines.}
Feature lines in garment design serve as critical elements for understanding garment construction, defining style characteristics, and analyzing pattern layouts. While recent research has made significant progress in garment generation with patterns~\cite{long2023neuraludf, yu2023surf, liu2024automatic, he2024dresscode, GarmentCode2023}, the specific challenge of extracting and representing garment feature lines for pattern understanding remains largely unexplored. 

Our framework uniquely addresses the challenges inherent in each of these structured representations by providing a unified approach to point distribution transformation, enabling the capture of both geometric and semantic relationships across different domains of 3D shape analysis.



\section{Methodology}
\label{sec:methods}

Given a reference point cloud $\mathbf{P}_{\text{ref}} = \{ p_{\text{ref}}^{i} \in \mathbb{R}^3 \mid i = 1, \dots, N \}$ sampled from a source surface distribution $p_S(\mathbf{x})$, our objective is to generate a corresponding set of target points $\mathbf{P}_{\text{target}} = \{ p_{\text{target}}^i \in \mathbb{R}^3 \mid i = 1, \dots, N \}$ that characterize structural features. For training, we establish correspondences between each reference point $p_{\text{ref}}^{i}$ in $\mathbf{P}_{\text{ref}}$ and its associated target point $p_{\text{target}}^{i}$ in the ground truth structure, defined through a matching function $\mathcal{F}: \mathbb{R}^3 \rightarrow \mathbb{R}^3$ that maps reference points to their nearest target points. Our diffusion model learns to transform points from an initial Gaussian noise distribution towards these target positions through a guided denoising process. Specifically, we maintain explicit associations between each point in the diffusion trajectory and its corresponding reference point, enabling the model to learn the underlying geometric relationships that govern the transformation from $p_S(\mathbf{x})$ to $p_T(\mathbf{x})$.

\subsection{DDPM revisited}

We first revisit DDPMs and diffusion models. In Denoising Diffusion Probabilistic Models (DDPM)~\cite{ho2020denoising}, the diffusion model is formulated as a pair of processes: a forward noising process and a reverse denoising process. The forward process gradually adds Gaussian noise to the original data $\mathbf{x}_0$ through a sequence of steps, defined by the transition probability $q\left(\mathbf{x}_t \mid \mathbf{x}_{t-1}\right) = \mathcal{N}\left(\mathbf{x}_t ; \sqrt{1 - \beta_t} \mathbf{x}_{t-1}, \beta_t \mathbf{I}\right)$, where $\beta_t$ represents the scheduled noise variance at each timestep, with values between 0 and 1.
In the reverse process, the diffusion model is trained to learn a denoising network parameterized by $\boldsymbol{\theta}$, which inverts the forward process to recover the original data. This reverse process is defined as $p_{\boldsymbol{\theta}}\left(\mathbf{x}_{t-1} \mid \mathbf{x}_t\right) = \mathcal{N}\left(\mathbf{x}_{t-1} ; \boldsymbol{\mu}_{\boldsymbol{\theta}}\left(\mathbf{x}_t, t\right), \sigma_t^2 \mathbf{I}\right)$, where the network $\boldsymbol{\mu}_{\boldsymbol{\theta}}$ predicts the mean of the denoised distribution. In this formulation, $x_t$ can be sampled at time step $t$ in the form of: 
$q\left(x_t \mid x_0\right):=\mathcal{N}\left(x_t ; \sqrt{\bar{\alpha}_t} x_0,\left(1-\bar{\alpha}_t\right) \mathbf{I}\right)$, thus we have:
\begin{equation}
    x_t:=\sqrt{\bar{\alpha}_t} x_0+\sqrt{1-\bar{\alpha}_t} \epsilon, \quad \text { where } \epsilon \sim \mathcal{N}(\mathbf{0}, \mathbf{I})
    \label{equ:forward_sample}
\end{equation}
where $\alpha_t:=1-\beta_t$ and $\bar{\alpha}_t:=$ $\prod_{s=1}^t \alpha_s$.



\subsection{Geometry distribution transformation}

Our objective is to use diffusion models to learn the transformation from an arbitrary surface point distribution to a semantically meaningful target point distribution. In this paper, we demonstrate the transformation of surface sample points into three typical distributions: mesh key points, inner skeletal points, and feature line points. Inspired by Geometry Distributions~\cite{biao2024geometry}, our method introduces a point-wise guided diffusion framework where each point in the diffusion process maintains an explicit correspondence with a sampled reference point from the source surface. Unlike previous methods where conditioning is performed with the reference point cloud as a whole, our explicit per-point correspondence provides a flexible geometric representation where any sampling density from the source surface yields valid transformations while preserving geometric fidelity. 





We formalize our geometric transformation framework as a probabilistic mapping between two distributions: a source distribution $p_S(\mathbf{x})$ that characterizes the underlying surface geometry, and a target geometry distribution $p_T(\mathbf{x})$ that represents structural features. Given the surface distribution $p_S(\mathbf{x})$, we sample a set of reference points $\mathbf{P}_{\text{ref}} \sim p_S(\mathbf{x})$, and use this point set to guide a diffusion denoiser $\boldsymbol{\epsilon}_\theta(\mathbf{x}_t, t, \mathbf{P}_{\text{ref}})$. 

Following the denoising diffusion probabilistic model framework~\cite{ho2020denoising}, we use the identical forward progress to corrupt the initial distribution. The reverse process learns a parametric model $\boldsymbol{\epsilon}_\theta(\mathbf{x}_t, t, \mathbf{P}_{\text{ref}})$ that predicts the noise components conditioned on both the current state $\mathbf{x}_t$ and the reference points $\mathbf{P}_{\text{ref}}$. This enables guided denoising through the conditional probability $p_\theta(\mathbf{x}_{t-1}|\mathbf{x}_t, \mathbf{P}_{\text{ref}})$. During inference, the process begins with $\mathbf{x}_T \sim \mathcal{N}(\mathbf{0}, \mathbf{I})$ and iteratively samples points that converge to the target distribution $p_T(\mathbf{x})$, while the conditioning on $\mathbf{P}_{\text{ref}}$ ensures preservation of geometric relationships with the source surface throughout the generation trajectory.

\subsection{Model design}

\noindent As shown in Fig.~\ref{fig:model}, 
our model architecture is designed to effectively learn the geometry distribution transform between source and target point distributions while maintaining explicit point-wise correspondences throughout the diffusion process. Our design integrates a diffusion transformer (DiT)~\cite{peebles2023scalable} framework with a Point Voxel CNN (PVCNN)~\cite{liu2019point} feature extractor, where the PVCNN enables robust point feature extraction and the Diffusion Transformer guides the denoising process. 
This design choice is motivated by two key requirements from our problem formulation: (1) the need to process and maintain point-wise relationships between reference points $\mathbf{P}_{\text{ref}}$ and their corresponding target positions, and (2) the necessity to condition the denoising process on these geometric relationships. Accordingly, PVCNN efficiently extracts per-point features that capture local geometric context from the reference point cloud, while DiT's transformer architecture enables effective modeling of the conditional probability $p_\theta(\mathbf{x}_{t-1}|\mathbf{x}_t, \mathbf{P}_{\text{ref}})$ through its self-attention mechanisms. This combination allows our model to learn the underlying geometric relationships that govern the transformation from $p_S(\mathbf{x})$ to $p_T(\mathbf{x})$ while preserving point-wise correspondences throughout the diffusion trajectory.

We start by extracting per-point features from the reference point cloud $\mathbf{P}_{\text{ref}}$ using PVCNN, resulting in a feature matrix $\mathbf{F}_{\text{ref}} \in \mathbb{R}^{N \times C}$, where $N$ represents the number of points and $C$ is the dimensionality of the features. These extracted features act as references for guiding the noised key points toward their target structral positions. 

A learnable positional embedding function $\mathbf{E}_{\text{pos}}: \mathbb{R}^3 \rightarrow \mathbb{R}^C$ is applied. It takes the $(x, y, z)$ coordinates of each point in $\mathbf{P}_{\text{ref}}$ and outputs a corresponding feature embedding for each individual point. The extracted per-point features $\mathbf{F}_{\text{ref}} \in \mathbb{R}^{N \times C}$, together with the positional embeddings $\mathbf{F}_{\text{pos}} = \mathbf{E}_{\text{pos}}(\mathbf{P}_{\text{ref}})$, are combined to form the input tokens $\mathbf{F}_{\text{in}} = \mathbf{F}_{\text{ref}} + \mathbf{F}_{\text{pos}}$ for the DiT.

Additionally, we incorporate a timestep embedding $\mathbf{F}_{\text{time}} \in \mathbb{R}^{C}$ as a conditioning signal for the DiT, consistent with the original DiT structure. 
Each DiT block consists of two main branches: a self-attention pathway and a feed-forward network (FFN) pathway. The self-attention branch enables global information exchange across all points, while the FFN branch processes point features independently. Both branches are augmented with adaptive layer normalization and scale-shift operations conditioned on timestep embeddings.


\subsection{Noise schedule}
The noise schedule in a diffusion model, represented by the sequence ${\beta_t}$, determines the Gaussian noise level added at each timestep in the forward process. The schedule is designed to progressively corrupt the original data distribution until it approximates a standard Gaussian distribution. In the original DDPM~\cite{ho2020denoising}, a simple linear schedule for $\beta_t$ was employed. 

The semantically meaningful points (e.g., key points, joints) are typically compact with fewer vertices, while the number of sampling points is relatively high. Therefore, the ideal scenario for the sampling points is to be densely clustered around the few vertices of the structured mesh. This imposes high demands on the convergence accuracy of the diffusion model.
In our experiments, however, we observe that conventional noise schedules result in ``fuzzy" outputs, where points fail to converge precisely on the target vertices. We identify the cause as the relatively high noise levels near the end of the diffusion process (at timesteps close to $t=0$), where generated vertices should ideally form tight clusters. Large noise scales at these later stages introduce excessive variance, preventing the generated points from converging accurately.

\begin{figure}
    \centering
    \includegraphics[width=0.8\linewidth]{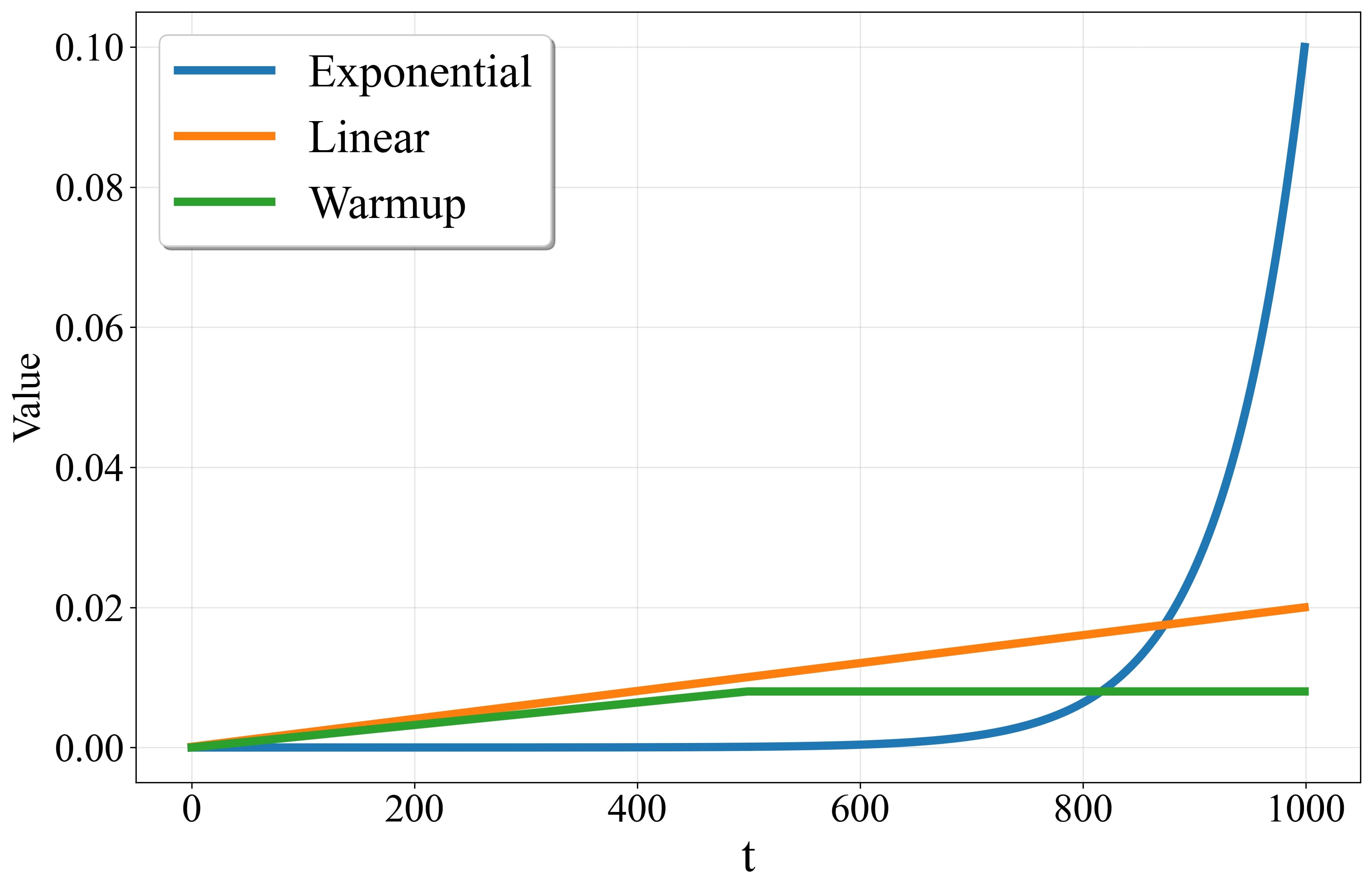}
    \caption{Comparison of different noise schedules. Our exponential schedule (blue) exhibits notably smaller $\beta_t$ values at lower timesteps, enabling more precise control during final denoising stages. 
    }
    \label{fig:noise_schedule_curves}
\end{figure}

To address this, we modified the noise schedule by setting the $\beta_t$ values close to $t=0$ to be significantly smaller. This adjustment reduces variance at critical points in the denoising process, allowing the generated vertex positions to converge more precisely. As shown in the accompanying graph Fig.~\ref{fig:noise_schedule_curves}, our modified schedule has smaller $\beta_t$ values at smaller $t$ steps, spending more effort on fine-grained details of our denoising process. 

Under this design, however, the value of $\beta_t$ close to $t=1000$ must be significantly larger to ensure proper training. Recall the sampling equation~\ref{equ:forward_sample}. To minimize the discrepancy between training and inference distributions, we need the variance $\sqrt{1-\bar{\alpha}_t}$ to approach 1 at large $t$, ensuring the forward process properly approximates a standard Gaussian distribution. We determined the terminal value to be $\beta_t = 0.1$, yielding $\sqrt{1-\bar{\alpha}_t} \approx 0.9998$, which empirically provides stable training while maintaining sufficient proximity to the standard Gaussian. Through experimentation, we find that our modified noise schedule robustly produces precise point clusters. 



\subsection{Sampling gradient adjustment}

Once the distribution transformation model is trained, during inference, we can impose constraints on the direction of distribution convergence to guide the target point distribution toward our desired outcome and thus enhance geometric precision, we introduce a sampling gradient adjustment (SGA) strategy that can be customized for different target distributions. 

The sampling process, detailed in Algorithm \ref{alg:Diffusion_Sampling}, introduces a correction gradient term at each denoising step. Inspired by ~\cite{chung2022diffusion}, for a given timestep $t$, we compute a geometry-aware gradient based on the current estimated positions $\hat{\mathbf{x}}_0$ and target geometric constraints. This gradient is then applied with a step size $\rho$ to guide points toward geometrically optimal positions. 



\begin{algorithm}
\caption{Sampling gradient adjustment for surface keypoints}
\label{alg:Diffusion_Sampling}
\begin{algorithmic}
\REQUIRE $T$, $\{\alpha_t\}_{t=1}^T$, $\{\sigma_t\}_{t=1}^T$, $\mathcal{M}$ , $\mathbf{P}_{\text{ref}}$, $\rho$
\STATE $\mathbf{x}_T \leftarrow \mathcal{N}(0, I)$
\FOR{$t = T$ down to $1$}
  \STATE $\mathbf{z} \leftarrow \mathcal{N}(0, I)$ if $t > 1$, else $\mathbf{z} = 0$
  \STATE $\hat{\mathbf{x}}_0 \leftarrow \frac{\mathbf{x}_t - \sqrt{1 - \bar{\alpha}_t} \, \boldsymbol{\epsilon}_\theta\left(\mathbf{x}_t, t\right)}{\sqrt{\bar{\alpha}_t}}$
  \STATE Find the closest point $\mathbf{c} \in \mathcal{M}$ to $\hat{\mathbf{x}}_0$
  \STATE Calculate adjustment gradient: $\nabla_{\mathbf{x}_t} = \|\hat{\mathbf{x}}_0 - \mathbf{c}\|_2^2$
  \STATE $\mathbf{x}_{t-1}' \leftarrow \frac{1}{\sqrt{\alpha_t}} \left( \mathbf{x}_t - \frac{1 - \alpha_t}{\sqrt{1 - \bar{\alpha}_t}} \, \boldsymbol{\epsilon}_\theta(\mathbf{x}_t, t, \mathbf{P}_{\text{ref}}) \right) + \sigma_t \mathbf{z}$
  \STATE $\mathbf{x}_{t-1} \leftarrow \mathbf{x}_{t-1}' - \rho \nabla_{\mathbf{x}_t}$
\ENDFOR
\RETURN $\mathbf{x}_0$
\end{algorithmic}
\end{algorithm}
We can impose different constraints based on the characteristics of the target distributions. For example, if the target distribution is surface-aligned (e.g., mesh keypoints), the gradient guides points toward the input surface $\mathcal{M}$, ensuring generated vertices maintain close surface alignment. The gradient is computed as the squared distance between each predicted point and its closest surface point. If the target point distribution is inside the surface (e.g., inner skeletal points), we can optionally use directional adjustment such as reverse normal direction to guide joints closer to the medial axis.
\section{Applications}
\label{sec:applications}
To demonstrate our method, we use three representative tasks where input points are transformed into distinct semantically meaningful geometric distributions: 1) surface-aligned mesh keypoints, 2) inner skeletal joints, and 3) continuous feature lines. These distributions highlight the varying structural relationships between the target and source point geometries. 

\subsection{Surface mesh keypoints}
The emergence of advanced 3D generation techniques has led to increasingly detailed and complex meshes. However, these high-resolution outputs often lack the carefully crafted structural properties that characterize artist-created meshes, making them less suitable for practical applications in animation, gaming, and interactive media. Our first application addresses this challenge by transforming dense surface samples into strategic vertex positions that guide mesh simplification toward artist-like results. Rather than solving remeshing directly, our focus is on generating semantically meaningful vertex proposals that guide classical simplification algorithms, bridging learning-based prediction with traditional geometry processing.

Given an input mesh $\mathcal{M}$, we first sample a dense set of reference points $\mathbf{P}_{\text{ref}}$ from its surface. Our PDT framework then transforms these points into a set of predicted vertex positions that capture key geometric features while maintaining the structural characteristics observed in artist-created meshes. These predicted positions are projected to their nearest surface triangle vertices and used as positional constraints in a modified Quadric Error Metrics (QEM) simplification process~\cite{garland1997surface}. 


\subsection{Inner skeletal joints}
Character animation in modern digital content creation relies heavily on skeletal rigs, which provide intuitive control mechanisms for artists to create natural movements and poses. While manual rigging remains common practice, the increasing demand for 3D character content necessitates automated solutions for skeleton generation. Our second application addresses this need by transforming surface geometry into meaningful skeletal joint positions.

Given an input character mesh $\mathcal{M}$, we employ our PDT framework to predict joint positions that reside within the shape volume. The framework processes surface samples $\mathbf{P}_{\text{ref}}$ from the input mesh, learning to transform these external points into internal skeletal joints that capture the character's anatomical structure. Our method learns to position joints at semantically meaningful locations, such as articulation points and medial regions.

\subsection{Continuous feature lines}
Feature line extraction from garment point clouds enables designers to understand garment construction, identify style elements, and analyze pattern layouts.

Given a dense point cloud sampling of a garment surface $\mathbf{P}_{\text{ref}}$, our framework transforms these input points into points on the continuous feature lines that delineate important structural elements such as stitches between patterns and boundaries. The PDT framework learns to identify and connect points that lie along meaningful feature paths, producing dense line approximations that capture both local details and global garment structure.

\section{Experiments}
\label{sec:experiments}

\subsection{Experiemental setups}
 We evaluate our PDT framework across three distinct tasks mentioned above: surface remeshing, skeletal joint prediction, and garment feature line extraction. We use standard DDPM with 1000 denoising steps. All generated keypoints are clustered with a simple thresholding.

\subsubsection{Remeshing.}
For the remeshing task, we train our diffusion model on the Objaverse dataset~\cite{deitke2023objaverse}, preserving original mesh structures to maintain artist-designed priors. We curate the dataset by filtering out meshes with more than 8,192 vertices or poor mesh quality, resulting in approximately $\sim$ 80k training meshes and $\sim$ 500 test meshes. During training, we uniformly sample 8,192 points from each input mesh surface. Our architecture employs a DiT model with 24 layers, 12 attention heads, and a hidden dimension of 768. The predicted keypoints serve as inputs to a constrained QEM algorithm~\cite{garland1997surface} for final mesh generation. The model is trained on 8 NVIDIA A6000 GPUs for $\sim$ 7 days.

\subsubsection{Skeletal Joints.}
For skeletal joint prediction, we utilize the ModelsResource RigNetv1 dataset~\cite{xu2020rignet, xu2019predicting}, which comprises diverse articulated 3D characters including humanoids and animals. The dataset contains 2,163 training samples and 270 test samples. We sample 2,048 points uniformly from each input character mesh. Our implementation uses a DiT model configured with 8 layers, 8 attention heads, and a hidden dimension of 512. The model is trained on 2 NVIDIA 3090 GPUs for $\sim$ 2 days.

\subsubsection{Garment feature lines.}
For garment feature line extraction, we leverage the GarmentCode dataset~\cite{GarmentCodeData:2024, GarmentCode2023}, which provides synthetic 3D made-to-measure garments with corresponding sewing patterns. The dataset features diverse garment categories fitted to various body shapes with different textile materials. We randomly sample a subset comprising 9,000 training samples and 1,000 test samples. Each garment is represented by 2,048 uniformly sampled surface points. The DiT model architecture consists of 8 layers, 8 attention heads, and a hidden dimension of 512. The model is trained on 2 NVIDIA 3090 GPUs for $\sim$ 2 days.

\subsection{Comparisons: remeshing}
We evaluate our model's effectiveness in mesh generation through comprehensive comparisons with state-of-the-art remeshing algorithms, including the Blender remeshing tool~\cite{blender}, Quadric Error Metrics (QEM)~\cite{garland1997surface}, CWF~\cite{xu2024cwf}, and Spectral remeshing~\cite{lescoat2020spectral}. All quantitative metrics are computed by uniformly sampling 8,192 points from both the input mesh and simplified mesh surfaces, and averaged over the test dataset.

\begin{figure*}
    \centering
    \includegraphics[width=0.7\linewidth]{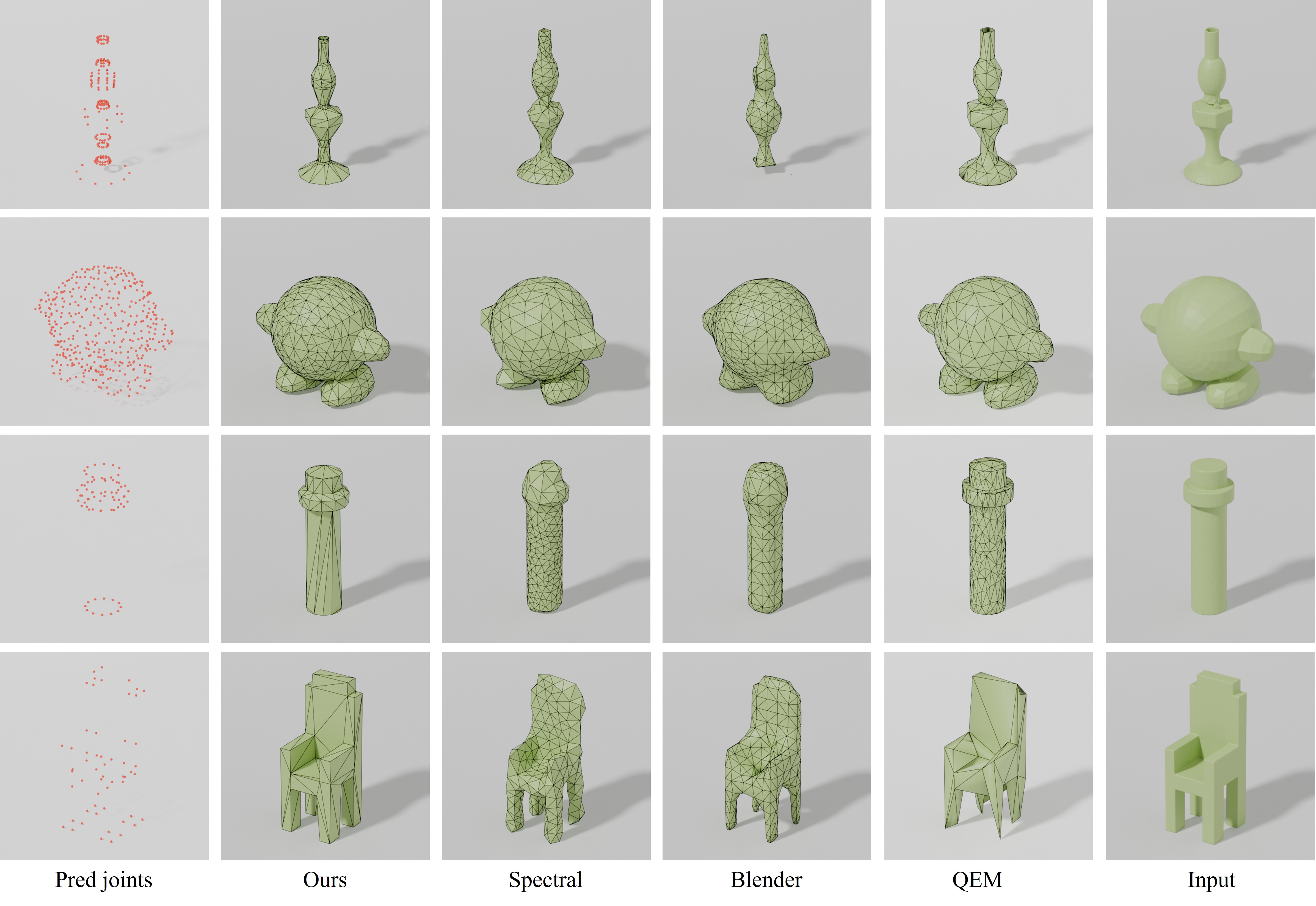}
    \caption{Visual comparison of remeshing results across different methods. The left-most column demonstrates our method's predicted vertex positions, which guide the subsequent mesh generation process. Note how our method preserves key structural features while maintaining regular mesh patterns.}
    \label{fig:comp_remesh}
\end{figure*}

\subsubsection{Qualitative analysis.}
Fig.~\ref{fig:comp_remesh} presents a visual comparison of remeshing results across different methods. The leftmost column shows the vertex positions predicted by our diffusion model, demonstrating how our approach strategically places vertices to capture key structural features. 
In contrast to baseline methods that often produce irregular mesh structures, our method generates more coherent and structurally-aligned meshes. 

\subsubsection{Quantitative evaluation.}
Table~\ref{tab:comp_remesh} presents a comprehensive quantitative comparison using standard geometric fidelity metrics: Chamfer Distance (CD), Maximum Mean Discrepancy (MMD) and Coverage (COV)~\cite{achlioptas2018learning}, and 1-Nearest Neighbor Accuracy (1-NNA)~\cite{lopez2016revisiting}. Additionally, we compare the mesh complexity in terms of vertex count (\#V) and face count (\#F). Our method achieves competitive metrics in all categories.

\begin{table}
    \centering
    \caption{Quantitative comparison with existing remeshing methods.}
    \small
    \setlength{\tabcolsep}{4pt}
    \begin{tabular}{|l|c|c|c|c|c|c|}
        \hline
        Method & CD($\times 10^{-2}$) $\downarrow$ & MMD($\times 10^{-2}$) $\downarrow$ & COV $\uparrow$ & 1-NNA $\downarrow$ & \#V & \#F \\
        \hline
        QEM &3.11 & 1.22 & \textbf{0.56} & \textbf{0.56} & 103 & 203 \\
        Blender & 5.62 & 1.39 & 0.47 & 0.83 & 275 & 543 \\
        Spectral & 4.50 & 1.20 & 0.51 & 0.74 & 269 & 546 \\
        CWF & 5.52 & 2.12 & 0.52 & 0.69 & 124 & {250} \\
        Ours & \textbf{2.82} & \textbf{1.10} & 0.53 & 0.72 & 106 & 201 \\
        \hline
    \end{tabular}
    \label{tab:comp_remesh}
\end{table}
We also conducted a user study with 24 participants comparing results from QEM, CWF, Spectral, and our method across 8 representative meshes. Participants rated each result based on sharp-feature preservation, artist-like mesh quality, and overall visual appeal. Our method was preferred in all aspects against existing methods.

\begin{table}
    \centering
    \caption{User study from 24 users in terms of sharp-feature preservation, artist-like mesh quality, and overall visual appeal. The number means the percentage of users favoring a specific method.}
    \small
    \setlength{\tabcolsep}{4pt}
    \begin{tabular}{|l|c|c|c|c|}
        \hline
        {Favourite (\%, $\uparrow$)} & {CWF} & {QEM} & {Spectral} & {Ours} \\
        \hline
        {Sharp Feature} & {14.1} & {15.1} & {1.0} & {\textbf{69.8}} \\
        {Artist-like} & {12.0} & {9.4} & {6.8} & {\textbf{77.1}} \\
        {Overall Visual} & {13.1} & {12.0} & {2.0} & {\textbf{74.3}} \\
        \hline
    \end{tabular}
    \label{table:user_study}
\end{table}



\subsection{Comparisons: skeletal joints}
We evaluate our method's performance in skeletal joints prediction against two state-of-the-art rigging methods: Pinocchio~\cite{baran2007automatic} and RigNet~\cite{xu2020rignet}.

We employ four complementary metrics to comprehensively assess joint prediction accuracy: CD-J2J measures the average bi-directional distance between each predicted joint and its nearest reference joint (and vice versa); Intersection over Union (IoU) quantifies skeleton similarity by computing the ratio of matched joints (using Hungarian matching) within a distance tolerance(in our case, 0.1) to the total number of joints; Precision (Prec) represents the fraction of predicted joints that match to reference joints within the tolerance, while Recall (Rec) indicates the fraction of reference joints that match to predicted joints within the same tolerance. As shown in Table~\ref{tab:comp_rig}, our method demonstrates superior performance across all evaluation criteria.

\begin{table}
    \centering
    \caption{Comparison of joint prediction accuracy across different methods. Our method achieves better performance in all categories.}
    \begin{tabular}{|l|c|c|c|c|}
        \hline
        Method & CD-J2J($\times 10^{-3}$)$\downarrow$ & IoU$\uparrow$ & Prec$\uparrow$ & Rec$\uparrow$ \\
        \hline
        Pinocchio & 9.01 & 53.0\% & 72.1\% & 64.6\% \\
        Rignet & 7.62 & 47.0\% & 70.4\% & 60.7\% \\
        Ours & \textbf{6.24} & \textbf{57.5}\% & \textbf{84.8}\% & \textbf{65.2}\% \\
        \hline
    \end{tabular}
    \label{tab:comp_rig}
\end{table}

\begin{figure}
    \centering
    \includegraphics[width=\linewidth]{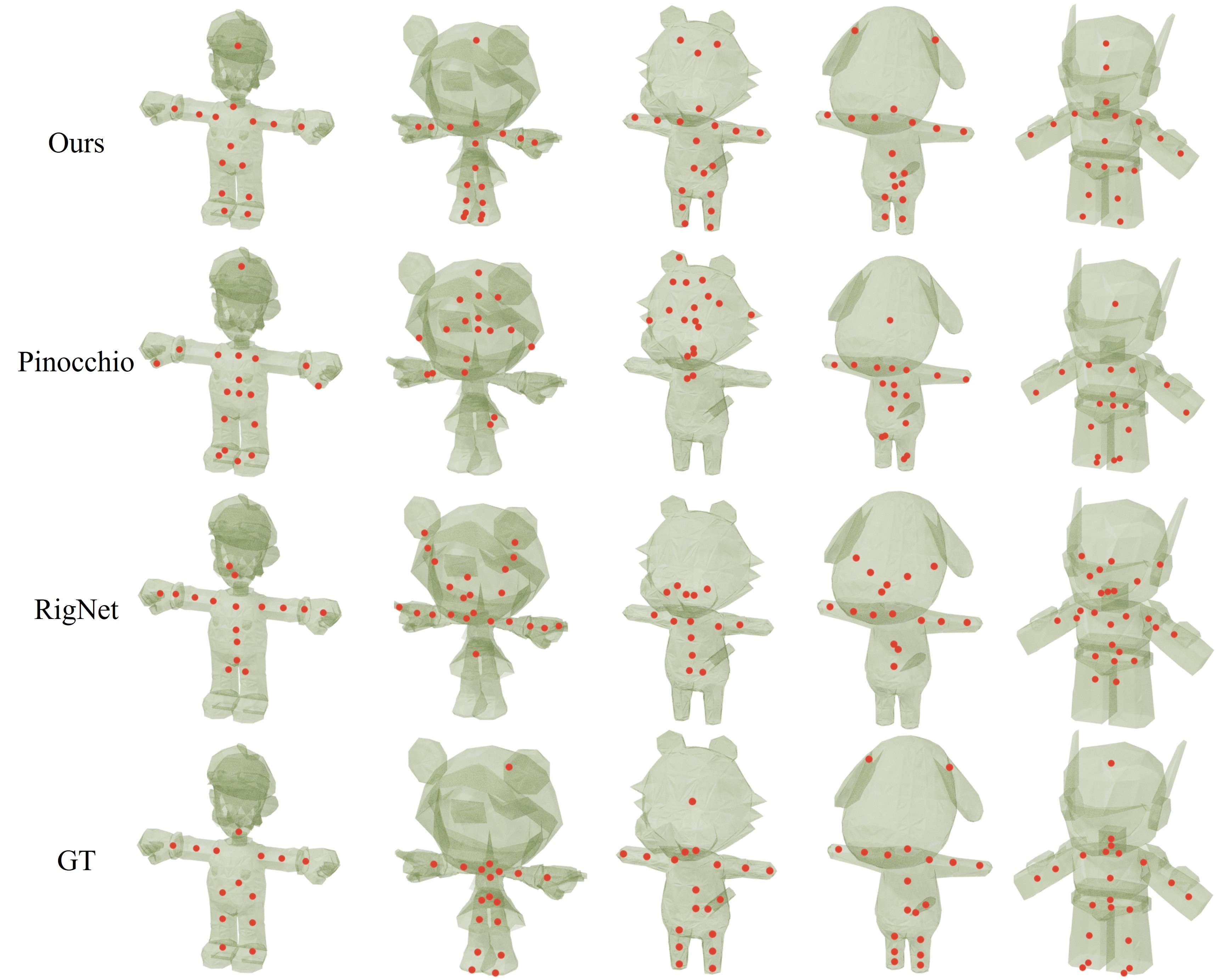}
    \caption{Visual comparisons between joint prediction results. Our method demonstrates better alignment with the underlying geometry and more anatomically plausible skeletal structures compared to baseline methods.}
    \label{fig:comp_joints}
\end{figure}

Fig.~\ref{fig:comp_joints} provides visual comparisons of joint prediction results across different methods. While Pinocchio~\cite{baran2007automatic} and RigNet~\cite{xu2020rignet} sometimes misses joint positions at critical positions such as ankles and knees, our method produces more complete skeleton structures with higher accuracy.

\subsection{Discussions}
\begin{figure}
    \centering
    \includegraphics[width=\linewidth]{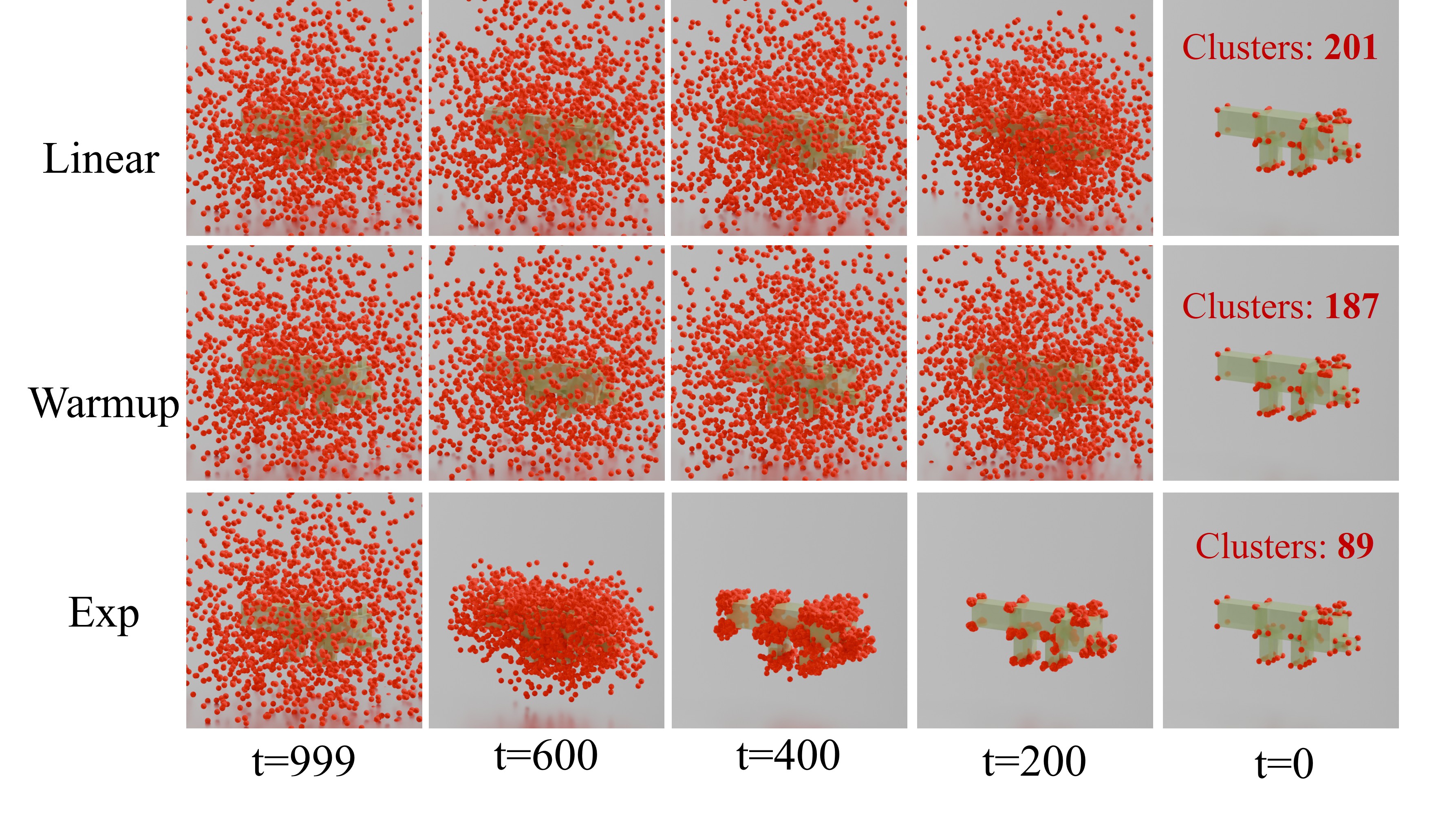}
    \caption{Visualization of the denoising progressions. The linear and warmup schedules exhibit abrupt transitions from noise to final output, while our exponential schedule demonstrates a gradual denoising process that yields tightly consolidated points with less clusters after thresholding.}
    \label{fig:denoise_prog}
\end{figure}

\subsubsection{Noise schedule.}
We conduct a comparative analysis of three distinct noise schedules: the linear schedule proposed in the original DDPM~\cite{ho2020denoising}, a warmup schedule where it starts slowly and then grows faster linearly, and our proposed exponential schedule. To evaluate their effectiveness for surface mesh keypoint generation, we train identical models under each schedule while maintaining all other setups constant.


Fig.~\ref{fig:denoise_prog} illustrates the progression of the denoising process under each schedule. The visualization reveals that both linear and warmup schedules produce "fuzzy" regions where points fail to form distinct, well-separated clusters at target locations. In contrast, our exponential schedule generates tightly clustered point distributions that precisely converge to target positions, achieving less number of clusters after thresholding. 


\subsubsection{Sampling gradient adjustment.}
We analyze the effectiveness of our sampling gradient adjustment (SGA) mechanism in improving the geometric accuracy of predicted points. 
\begin{figure}
    \centering
    \includegraphics[width=\linewidth]{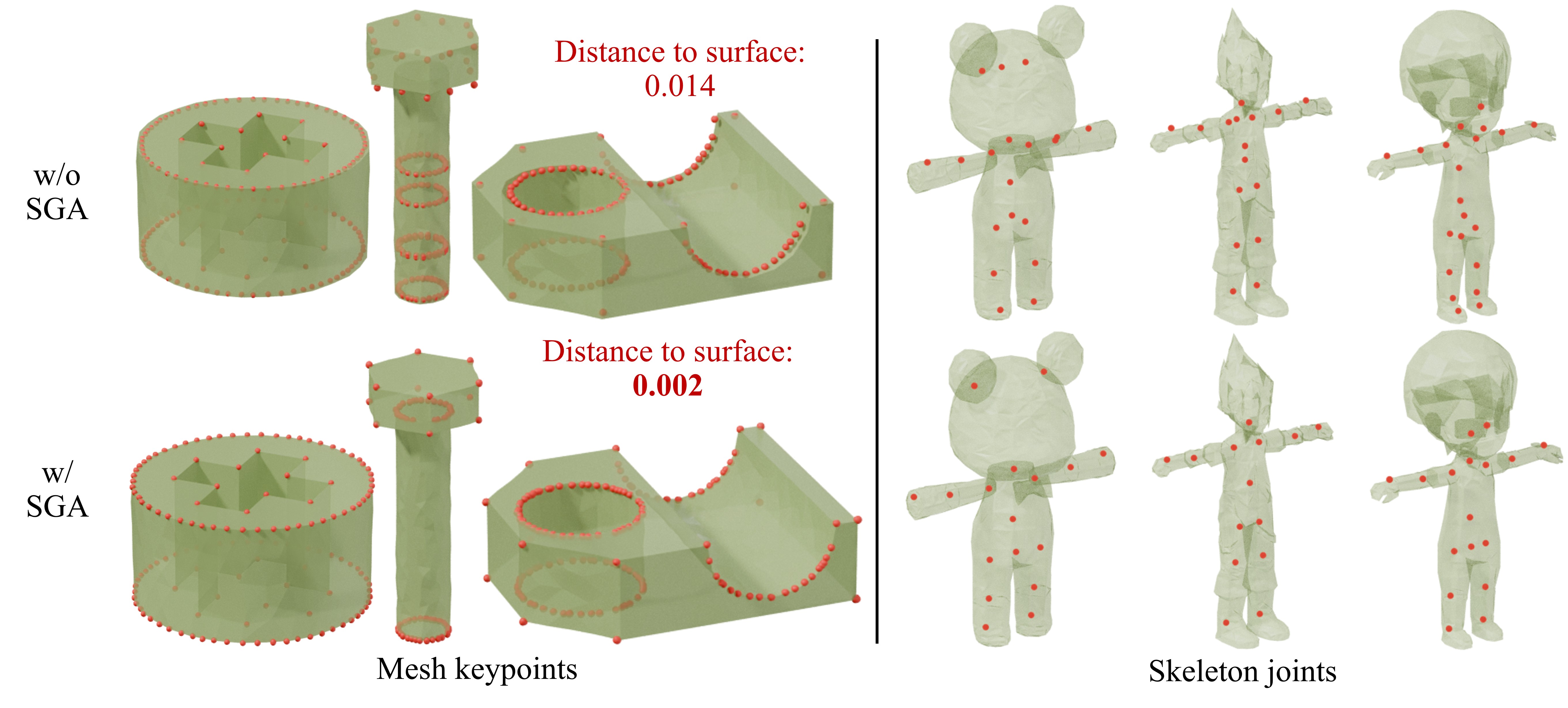}
    \caption{Impact of sampling gradient adjustment (SGA) on point placement accuracy. Left: Surface keypoint predictions with and without SGA, showing improved surface alignment. Right: Skeletal joint predictions, demonstrating enhanced medial positioning within the shape volume.}
    \label{fig:comp_dps}
\end{figure}
Without SGA, predicted keypoints exhibit noticeable deviation from the mesh surface. With SGA enabled, we observe consistently tighter surface adherence, with lower distance to surface as shown in Fig.~\ref{fig:comp_dps} (left). skeletal joints predicted with SGA are also more accurately positioned at the medial axes of the shape, better reflecting the natural skeletal structure of the input shapes, as showcased in Fig.~\ref{fig:comp_dps} (right).

\subsubsection{Feedforward method.}
We compare our method against a PVCNN-based feedforward model~\cite{liu2019point} with more number of parameters. For skeletal joints prediction, the feedforward baseline produces an average of 56.9 joints, compared to the ground-truth average of 20.8. Their mean IoU and precision scores dropped significantly (20.0 \% / 45.8\% respectively). In the remeshing task, PVCNN overpredict keypoints by 700 points on average, making the outputs unsuitable for constraint-based simplification. These results underscore the importance of our point-wise guided diffusion process and exponential noise schedule for generating sparse, semantically aligned structures.

\subsubsection{Sampling points number.}
\begin{figure}
    \centering
    \includegraphics[width=\linewidth]{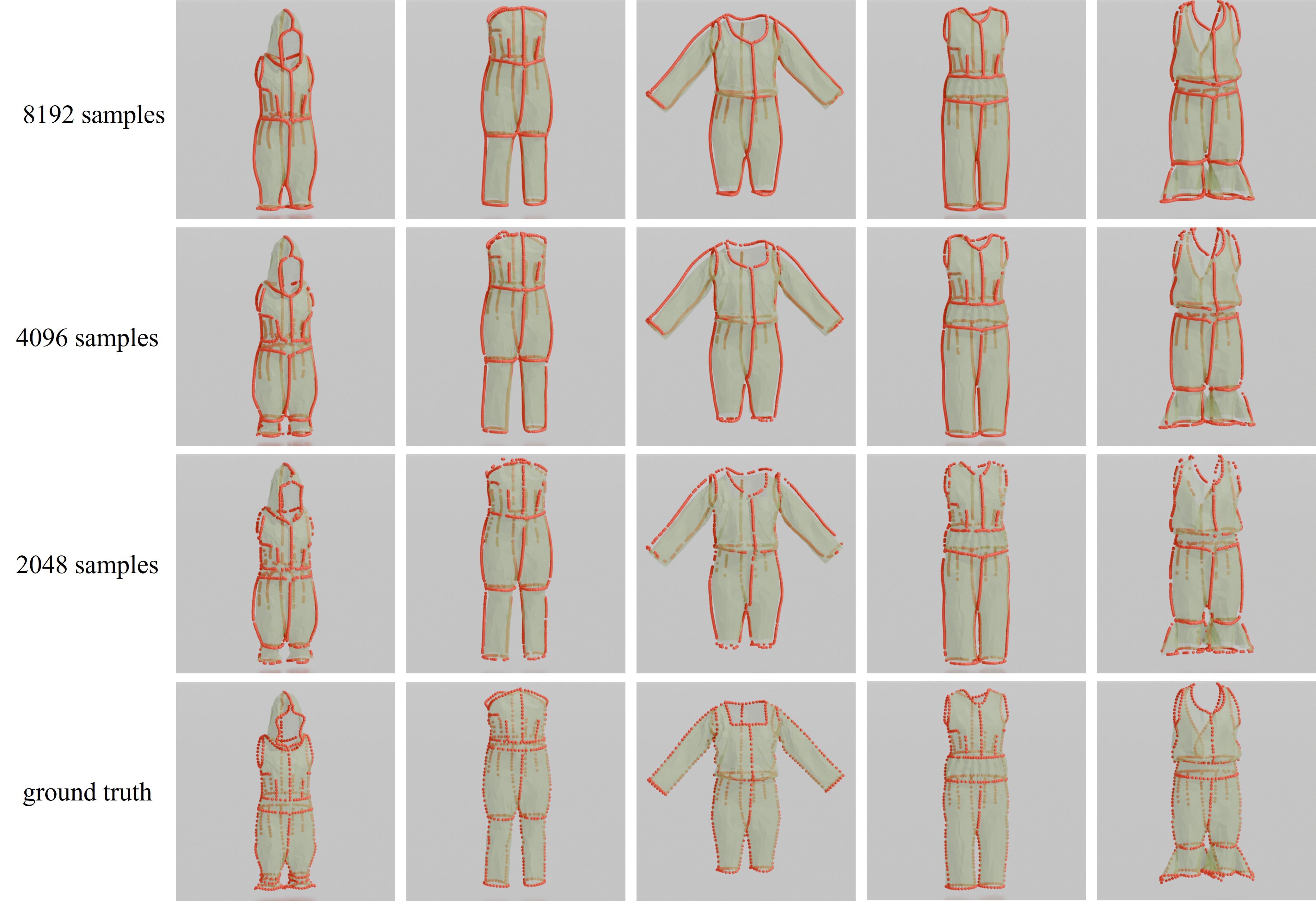}
    \caption{Feature line extraction results with varying input sample densities. From left to right: results using N=2048, 4096, and 8192 surface points. }
    \label{fig:comp_garments}
\end{figure}

While our model is trained on ground truth data containing 2048 sparse feature line points, it demonstrates effective generalization to higher sampling points. As shown in Fig.~\ref{fig:comp_garments}, PDT successfully learns to map surface points to their nearest target feature line positions, maintaining consistent quality even with significantly denser inputs than seen during training. 

\section{Conclusions}
We propose PDT, a novel framework that employs diffusion models for point distribution transformation. Given an input set of points, PDT learns to transform from the original geometric distribution to a semantically meaningful target distribution. The framework’s unique architecture and a set of novel designs enable diffusion models to effectively link the source and target point distributions through a denoising process. Our method is validated through three tasks, each involving the transformation of input points into distinct semantically meaningful geometric distributions: 1) surface-aligned mesh keypoints, 2) inner skeletal joints, and 3) continuous feature lines. Extensive experiments demonstrate that our method excels at generating diverse structurally meaningful point distributions through learned transformations. The proposed framework serves as a versatile solution, capable of being applied to a wide range of point cloud distribution learning and transformation tasks.

\label{sec:conclusion}

\begin{acks}
We would like to thank the anonymous reviewers, committee members for their insightful comments and suggestions. Jionghao Wang, Wenping Wang and Xin Li are partially supported by National Science Foundation (CBET-2115405). Rui Xu, Zhiyang Dou and Taku Komura are partially supported by the Research Grants Council of Hong Kong (Ref: 17210222), the Innovation and Technology Commission of the HKSAR Government under the ITSP-Platform grants (Ref: ITS/319/21FP, ITS/335/23FP), and the InnoHK initiative (TransGP project). Rui Xu, Zhiyang Dou and Taku Komura conduct part of the research in the JC STEM Lab of Robotics for Soft Materials, funded by The Hong Kong Jockey Club Charities Trust.
\end{acks}

\bibliographystyle{ACM-Reference-Format}
\bibliography{main}

\newpage

\appendix

\begin{figure*}
    \centering
    \includegraphics[width=0.95\linewidth]{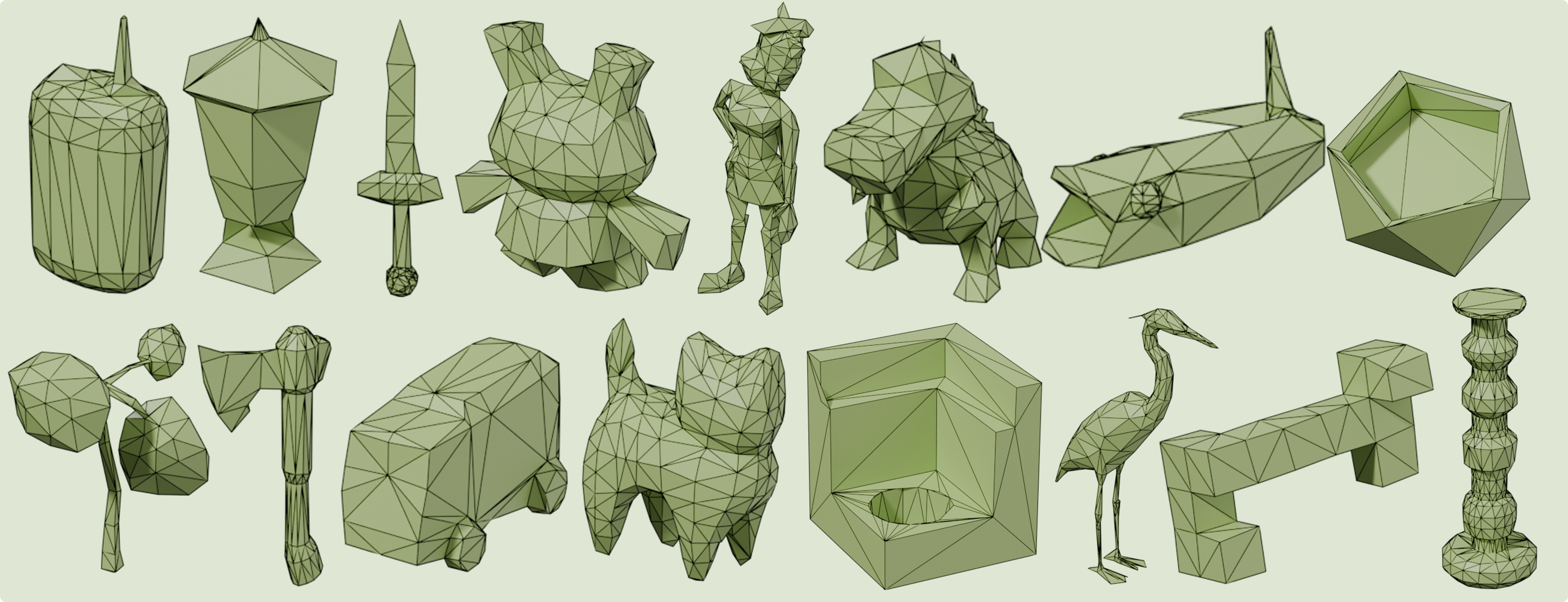}
    \caption{Gallery of our remeshing results.}
    \label{fig:gallery_remesh}
\end{figure*}

\begin{figure*}
    \centering
    \includegraphics[width=0.95\linewidth]{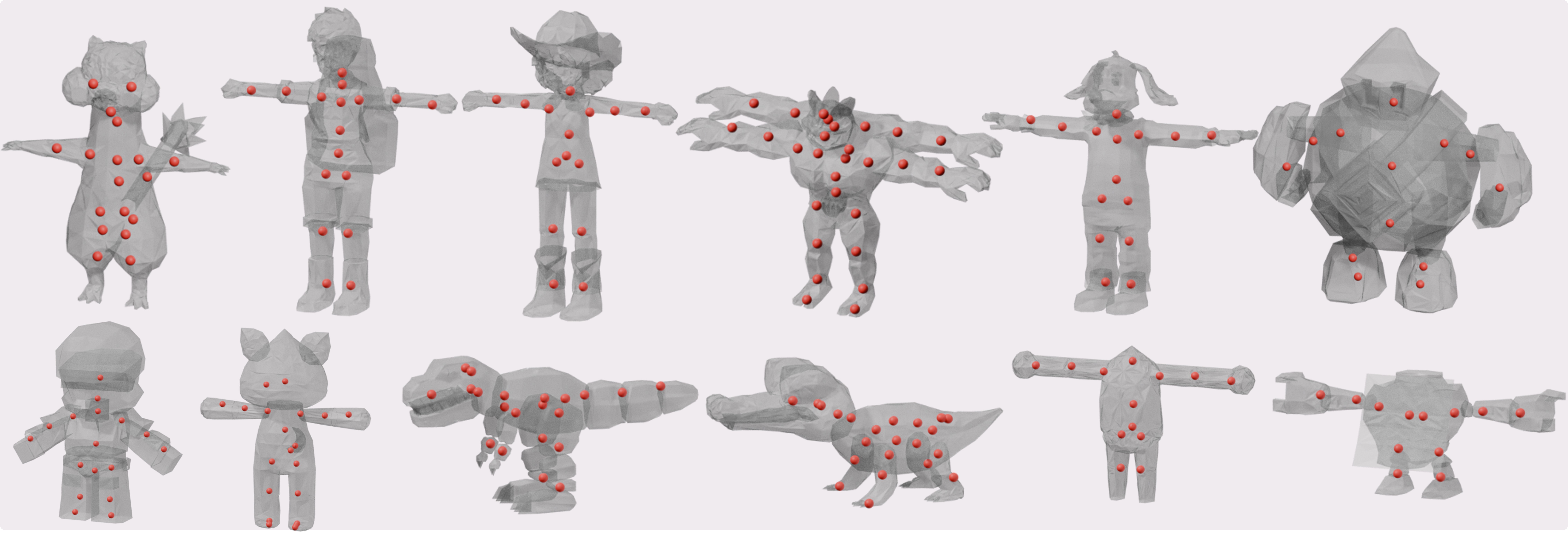}
    \caption{Gallery of our skeletal joints results.}
    \label{fig:gallery_joints}
\end{figure*}

\begin{figure*}
    \centering
    \includegraphics[width=0.95\linewidth]{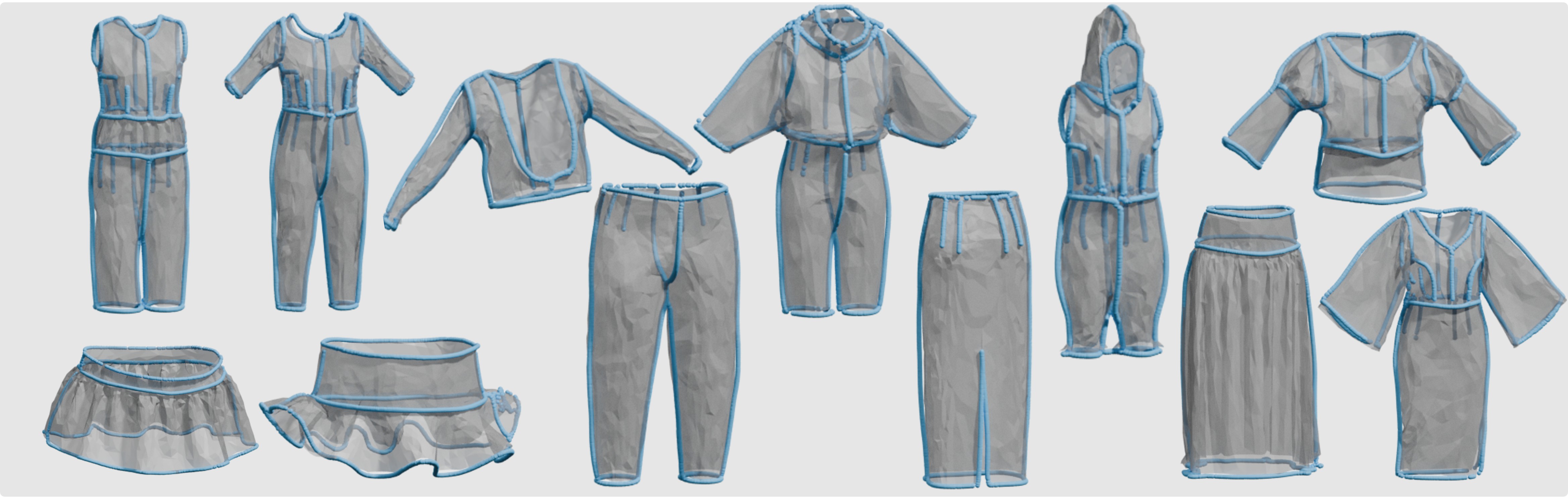}
    \caption{Gallery of our garment feature line extraction results.}
    \label{fig:gallery_garments}
\end{figure*}


\end{document}